# <tiger2/> – Serialising the ISO SynAF Syntactic Object Model

Laurent Romary (laurent.romary@inria.fr)[1,2], Amir Zeldes (amir.zeldes@georgetown.edu)[3] and Florian Zipser (f.zipser@gmx.de)[1,2]

[1] Inria
[2] Humboldt-Universität zu Berlin, Institut für deutsche Sprache und Linguistik
[3] Georgetown University

**Abstract**
This paper introduces <tiger2/>, an XML format developed to serialise the object model defined by the ISO Syntactic Annotation Framework SynAF. Based on widespread best practices we adapt a popular XML format for syntactic annotation, TigerXML, with additional features to support a variety of syntactic phenomena including constituent and dependency structures, binding, and different node types such as compounds or empty elements. We also define interfaces to other formats and standards including the Morpho-syntactic Annotation Framework MAF and the ISOCat Data Category Registry. Finally a case study of the German Treebank TueBa-D/Z is presented, showcasing the handling of constituent structures, topological fields and coreference annotation in tandem.

## 1 Introduction

The provision of reliable syntactic annotation for written or spoken linguistic data represents an essential step for a variety of natural language processing tasks, as well as for the understanding of more content-related semantic or pragmatic linguistic phenomena. In order to maximise the usefulness of such annotations it is essential that standardised representations for the interchange of syntactic data be defined and widely used within the linguistic and computational linguistic community.

At present the situation appears to be particularly complex because of the variety of contexts and forms that syntactic information may take. Firstly, syntactic information may either be the result of an automatic parsing of textual data or may be manually generated as a component of an annotated corpus. Secondly, the organisation and actual complexity of syntactic information highly depends on both the application context and theoretical background of the project within which such data has been created. In the simplest cases, syntactic representations may boil down to the identification of structural chunks on the textual surface, which may lead for instance to the further identification of technical terms, named entities or conceptual relations (Maedche & Staab 2000) in large quantities of data. By contrast, the validation of specific theoretical frameworks requires in-depth syntactic representations, often using a smaller sample of sentences, to account for the capacity of the corresponding theory to adequately represent some complex phenomena.

In between, comparatively large constituent treebanks (e.g. the Penn Treebank, Bies et al. 1995, TueBa-D/Z, Telljohann et al. 2004, 2009) and dependency treebanks (e.g. the Prague Dependency Treebank, Hajič et al. 2006) have been annotated with syntactic information following a more or less theory independent perspective, in order to offer a reference point for a variety of linguistic studies or for the development and testing of syntactic parsers. The complexity of the standardisation task arises from the need to uniformly represent data in a way that can do justice to all varieties of information from very disparate projects, retaining features from accepted best practices and without becoming unwieldy. Recent years have seen a wide range of standardisation initiatives which attempt to standardise models and formats of more or less particular subdomains in the linguistic area, such as MAF for morpho-syntactic information (ISO/DIS 24611; see also Romary & Witt, 2012) or more generally LAF, the Linguistic Annotation Framework (ISO/DIS 24612; see Romary & Ide, 2004 and Ide & Suderman, 2014), as well as more general efforts within the Text Encoding Initiative (TEI, see Burnard & Bauman 2008) and generalised XML formats such as GrAF (Ide & Suderman 2007). For syntax, the syntactic annotation framework SynAF (ISO 24615, see Bosch et alii, 2012) is the ISO standard defining a general data-model for syntactic objects and basic inventories with which syntactically annotated resources can be constructed. The goal of the present article is to introduce an XML serialisation based on TigerXML (Mengel & Lezius 2000) for concrete realisations of the SynAF data-model which follows accepted best practices in the treebanking community, is adequately powerful for its needs and avoids unnecessary complications.

## 2 Models and standards for language resources

As alluded to in the introduction, the <tiger2/> initiative is part of a wider endeavor to provide the linguistic resource domain with a complete portfolio of standards facilitating the interchange of linguistic data or the interconnection of language processing tools. In the last 25 years, such initiatives have mainly taken place within the TEI community, with the delivery and maintenance of the TEI guidelines, but also within the International Organization for Standardization (ISO), where its technical committee TC 37/SC 4 has already published several essential standards, with currently even more on its work plan. In the following paragraphs we will try to provide a quick overview of these initiatives with the perspective of better understanding the relationship between concrete formats and the underlying data-models.

The reference example we may begin with is the ISO standard 24610-1 (see Lee et al., 2004) for the XML representation of feature structures issued in 2006. This work, the first published by ISO committee TC 37/SC 4, resulted from the conjunction of two favorable factors. On the one hand, the theoretical linguistics domain and in particular several syntactic theories had long since identified the formal background for modeling typed feature structures (cf. Pollard & Sag 1994). On the other hand, feature structures have been widely used for even very simple annotation tasks in linguistics (e.g. in phonetics) and this lead the TEI to design an early SGML (then XML) format for their representation (Langendoen & Simons 1995). As a consequence, it was a straightforward move that lead to the resulting ISO standard, which took up most of the TEI components and expanded it to be fully compliant with the general model of typed feature structures.

In the case of ISO 24613 (LMF, see Romary 2013a), published in 2008, there was at that time no real generic model for the representation of lexical data, but a variety of more or less stabilized formats (see for instance Ide & Véronis 1995 and Romary, 2013a for an overview). The standard thus focused on providing a flexible meta-model (with extensions) encompassing a variety of lexical forms (morpho-syntactic, syntactic-lexical, machine readable dictionaries, etc.) and paying reduced attention to defining a comprehensive serialization for it. Work thus remains to be done in this respect and as shown in (Romary 2013b) it may be an opportunity to achieve a better convergence with large coverage vocabularies such as the TEI guidelines.

Finally, it can often be argued that the variety of models and formats that are related to the representation of language resources may be a hindrance to a real interoperability of linguistic data. Whereas this is typically a topic covered by GrAF for integrated language resources or PAULA (Dipper 2005) for the back-office representation within a corpus management environment such as ANNIS (Zeldes et al. 2009), we consider the main issue to be preserving the semantics of the underlying models as well as those of elementary descriptors that are combined with them, in the spirit of ISO 12620 (Data Category Registry/ISOCat, see below). In this respect, well defined language resources, which follow the ISO principles applied in this paper, can be disseminated further according to the Linked Open Data principles (see http://linkeddata.org/ and the Linguistic Linked Open Data initiative: http://linguistics.okfn.org/resources/llod/).

In this landscape, we will see that the definition of <Tiger2/> is indeed very close to our first example with feature structures in that an existing model (SynAF), combined with an already well recognized serialization (TigerXML) is a good candidate for a reference standard in its domain.

## 3 The ISO SynAF initiative

The SynAF standard was initiated within ISO committee TC 37/SC 4 in 2006 with the support of the European project Lirics[1]. Committee TC 37/SC 4 was put together in 2002 within ISO to cover the necessary standardisation activities in the domain of language resources, and in particular to provide, step by step, a comprehensive portfolio of standards improving interoperability across Language Technology applications. In this respect, SynAF was conceived as compliant with the main design principles within ISO/TC 37/SC 4 right from the outset, i.e.:

- The necessity to rely on a generic modelling scheme to account for the variety of potential applications of syntactic representation

---

[1] http://lirics.loria.fr/

- To be articulated with the ISO Data Category Registry (DCR), as described in ISO 12620, and linked to the ISOCat platform.
- To be compatible with further standards development within ISO/TC 37/SC 4, in particular in the domains of multi-layered stand-off annotation (LAF, ISO/DIS 24612; see also Romary & Ide 2004) and morpho-syntactic annotation (MAF, ISO/DIS 24611)

From a modelling point of view, ISO standards and initiatives in the domain of language resources rely on a modelling framework (cf. Romary 2001) that construes the description of a linguistic representation (annotation scheme or lexical structure) as the combination of two main components:

- A *meta-model* that informs the general characteristics of the corresponding family of formats, which is described as the combination of elementary representational units (components).
- Data categories, which represent elementary linguistic properties (or values thereof) attached to the various components of the meta-model.

A specific combination of a meta-model with a data category selection provides a full specification of all interoperable formats for a given type of linguistic annotation or linguistic database.

The core part of the SynAF meta-model is based on the single notion of *syntactic node* — derived from the generic annotation node in LAF (Ide & Romary 2003). A syntactic node represents the elementary unit of syntactic information. Syntactic nodes are in turn connected with one another by means of *syntactic edges*, thus forming a *syntactic graph*. Both syntactic nodes and syntactic edges can be further constrained by annotations, i.e. elementary feature-value pairs expressing properties attached to them. As in all ISO compliant linguistic models, both attribute names and values of annotations should be defined in relation to standardised data categories in ISOCat or at least by providing an ISO 12620 compliant specification recorded in ISOCat. Although the present article is not concerned with particular values, we also integrate a mechanism for referencing ISOCat entries as part of the XML serialization of SynAF

To provide a better account of the various forms of syntactic annotation, the SynAF meta-model further refines the notion of syntactic node by providing two more specific classes, namely *terminal* and *non-terminal nodes*. Non-terminal nodes account for the abstract syntactic structures that have no direct anchoring on the surface of the language data to be annotated. These usually serve to represent syntactic constituents such as nominal and verbal phrases. Terminal nodes, on the contrary, are the places of articulation between syntactic structures and the linguistic data. Terminal nodes too can form independent graphs without non-terminal nodes, for example in the case of dependency trees or some forms of coreference annotation. Terminal nodes are also understood as instances of the *word form* class in ISO MAF, which can flexibly represent a variety of surface forms analysed as having the status of a word unit (including blends, acronyms, compounds, truncated forms and much more). This allows one to deploy various implementation strategies of MAF and SynAF depending on available data or theoretical framework (see below on MAF integration with SynAF in <tiger2/>). In the following paragraphs, we outline several such possible strategies.

**Fig. 1** Graphic representation of the SynAF meta-model in ECore, a UML near dialect (see Steinberg et al. 2009).

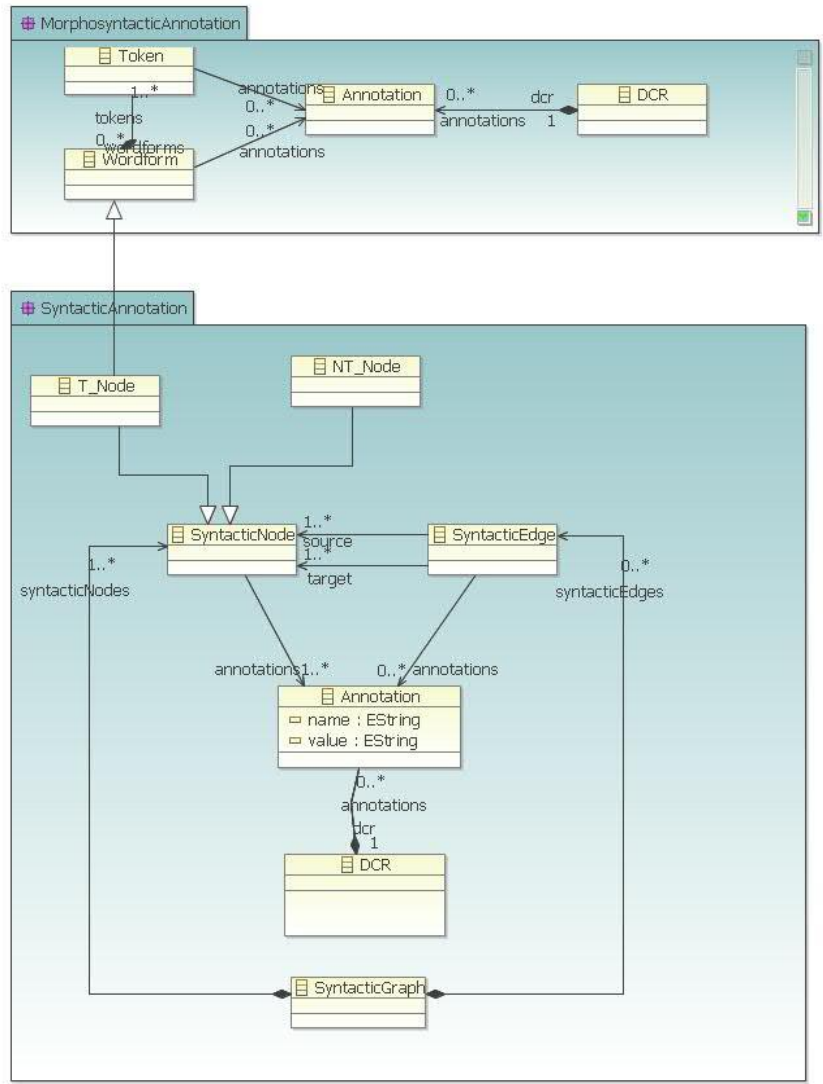

The meta-model in Figure 1 shows syntactic types in the bottom box, with elements for terminal and non-terminal nodes (T_Node and NT_Node), which are instances of the type SyntacticNode. Objects of these types are connected by syntactic edges (objects of the type SyntacticEdge), both of which are annotatable with annotations possibly connected to the DCR. The resulting graph can be connected to external models such as morpho-syntactic annotations through links between terminal objects and external objects as shown above. With this general architecture in mind, the exact expression of the model elements above in the more specifically articulated model of <tiger2/> is laid out below.

## 4  Serialising SynAF

Rather than further increase the diversity of already existing formats, we decided to serialise the SynAF meta-model by adapting an existing popular format for syntax annotation and changing only as much as required and as little as necessary. The choice of TigerXML as a basis for the serialisation of SynAF was primarily motivated by its reception in a wide community of computational and corpus linguists, its readability and its flexibility, as in the ability to define custom node and edge annotations in its declaration block, a mechanism we will extend below.

### 4.1  Extending the TigerXML data-model

While TigerXML has been very successful as a linguistic theory-neutral, versatile format for constituent-based annotations, its expressive power is limited by design decisions made in the 1990s in order to meet the needs of specific corpus annotation schemes, and some of its general features are now out-of-date (e.g. use of the deprecated `@idref` attribute and the lack of modern `@xml:id` attributes and XPointers). The possibility of typing nodes in Tiger XML was, similarly to SynAF, limited to two types, called non-terminal and terminal nodes. The mechanism of typing edges was also restricted to two types: primary and secondary dominance edges, whose labels can be freely specified in `@label`. This restricts the power of the TigerXML format to the annotation of only a subset of the syntactic phenomena which are representable with just one type of terminal or non-terminal node and only two types of edges; as we shall see below, these may not suffice for a variety of purposes, and there is no reason not to extend the inventory of types to an arbitrary size to meet users' needs. Further deficits which should be addressed from the contemporary perspective involve the ability to link and reference resources outside the syntactic annotation graph proper, including separately stored stand-off source text (for example to preserve whitespace, which the TigerXML tokenisation cannot represent), referencing external standards for annotation schemes using state-of-the-art repositories such as the ISOCat Data Category Registry (ISO 12620), and interoperable binding of annotation layers from other standards, such as the Morpho-syntactic Annotation Framework MAF for the annotation of morphological phenomena (ISO/DIS 24611).

#### 4.1.1  A meta-model approach

As with SynAF, <tiger2/> follows the design principles mentioned in Section 2 by implementing an explicit object model for its representational elements. Our approach is therefore not limited to providing a new XML-based format for syntactic data as described below, but rather also provides a meta-model following the model-based approach (see MDA, Miller & Mukerji 2003). Building on the <tiger2/> meta-model, we have created a Java API for <tiger2/> with EMF, the Eclipse Modeling Framework (Steinberg et al. 2009). We have used EMF in combination with the Eclipse IDE for this purpose, because they are open source, available across platforms and have good support for generating Java code. EMF also distinguishes a model layer and a persistencing layer. This distinction allows us to bind several formats to the same meta-model. More specifically we created a mapping between the TigerXML format and <tiger2/> as well as the <tiger2/> meta-model. Thanks to this mechanism, it is possible to convert data from the TigerXML format to <tiger2/> and vice versa via a Java API. After importing data from the persistence layer into the model layer, the API allows us to programmatically manipulate, transform and merge data from multiple sources. This API can be used as a basis for processing SynAF data and as a library within other tools. The benefit of using such an API is that the programmer does not have to deal with the persistencing layer anymore, but just works with an abstraction above the concrete data level. Even if the underlying data format changes, for instance when serializing a <tiger2/> model in JSON instead of XML, programmers do not need to concern themselves with differences between formats.

In the last few years, we have seen a trend to not only observe linguistic phenomena on isolated levels of annotation, but also to bring annotation levels together in new multi-layer corpora. Therefore new tools and formats, which are able to deal with multi-layer data have gained prominence. In many cases users may wish to extend existing resources from various tools and formats with syntactic annotations in <tiger2/>. This means that users have to recreate their existing corpus in parts using a variety of serializations. Imagine a corpus, which already contains a tokenization and morphosyntactic layers (like lemma and POS annotations) next to the primary data. Such pre-processed data can also be a basis for a new syntactic annotation layer. Therefore the re-tokenization and re-annotation of data could become a nuisance and is in any case error prone. Because of this, the conversion of reusable data becomes an important issue. But, when establishing a new format the conversion of existing formats into a new one such as <tiger2/>, and vice versa, is less realistic the larger the set of already existing formats is. Though we have already implemented a conversion between the TigerXML format and the new <tiger2/> format using the <tiger2/> API, a general conversion capability with a variety of formats is desirable. We therefore extend the converter framework Pepper (Zipser & Romary 2010). Pepper is a pluggable framework basing on OSGi (see: www.osgi.org/) used to convert data between a variety of linguistic formats. The Pepper framework uses the meta-model Salt (see Zipser 2009 for details) as an intermediate representation of any supported linguistic format. Mapping data into another format via a common intermediate model allows us to reduce the number of mappings between all possible pairs of $n$ formats from $n^2$-$n$ to $2n$ mappings. Like GraF, Salt is based on a graph model, allowing a very generic treatment of linguistic data. In contrast to GrAF, Salt acts on the model layer in memory and not as a persistent XML format.

The aim of Pepper is to provide an interface for programmers to create mappings between data in any linguistic format and Salt. As mentioned above, the extensibility of Pepper allows us to create a mapping between Salt and our <tiger2/> meta-model and also to plug that mapping into the framework as a Pepper module. The mapping consists of two modules, one to map Salt data to <tiger2/> data and one for the other way around. Figure 2 gives an overview of the architecture of Pepper and its pluggable nature. To implement the mapping, we used the previously described <tiger2/> Java API. The pepper module specifies a programatic mapping between two meta-models, also written in Java. Since Salt is very generic and defines a set of nodes, edges, labels and layers adapted to generalized linguistic meanings, it is not fixed to syntactic annotations only and can represent morphology, coreference, information structure and many other phenomena. This is turn allows us to convert a wide variety of existing resources in different formats to <tiger2/>, including GrAF, as well as to export <tiger2/> projects without having to write separate converters between each pair of formats. New Pepper mapping modules can be written in Java, Python, XSLT or QVT to convert further formats to <tiger2/> (in our case we have opted for Java mappers, which are more efficient than e.g. XSLT).

**Fig. 2** An architecture overview of the Pepper framework and its correspondence to Salt, and the plugin mechanism for new modules called mappers.

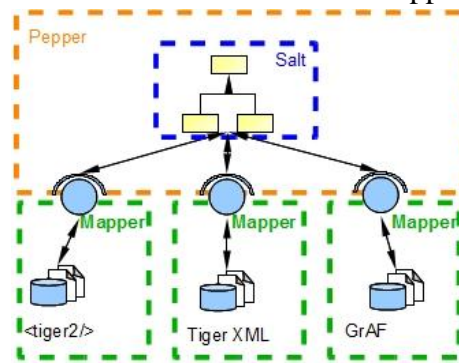

With a growing pluriverse of formats, the conversion between these formats becomes more and more important. On the one hand it is important to be backwards compatible and continue support for existing formats and tools that are no longer developed. On the other hand it is also important that a format like <tiger2/> be able to deal with further developments in the future.

### 4.1.2 The <tiger2/> meta-model and XML serialisation

Figure 3 shows the meta-model of <tiger2/> in Ecore, a UML-near syntax, generated with the Eclipse IDE and EMF. The meta-model of <tiger2/> derives from the meta-model of SynAF described above and is therefore fully compatible with the SynAF standard.

**Fig. 3** Graphic representation of the <tiger2/> meta-model in ECore.

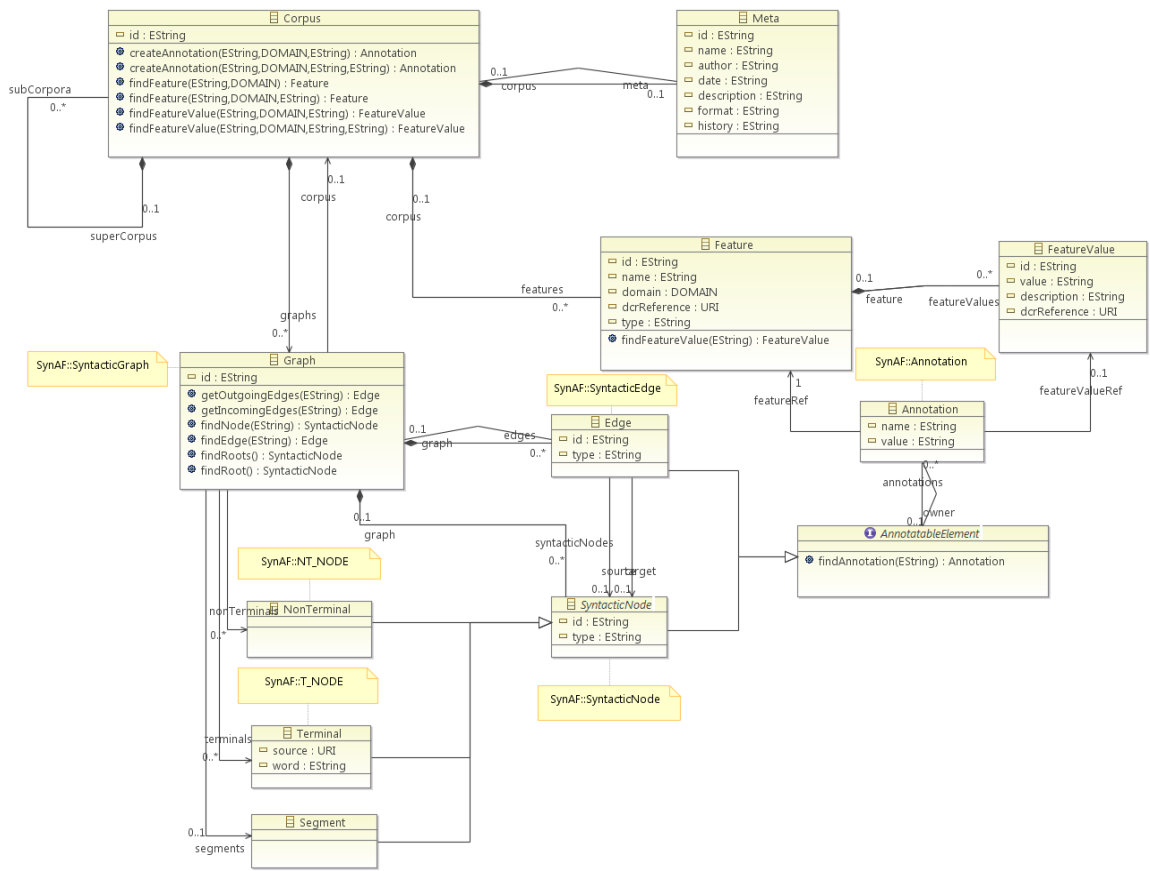

The elements describing the syntactic structure derived from the SynAF meta-model are *Graph*, *Edge*, *Terminal* and *Nonterminal*. In most cases, these elements will form a tree or DAG, but a graph including cycles is possible as well. The model elements *Terminal* and *Nonterminal* derive from *Node*. Together with *Edge* both are derived from the abstract element *AnnotatableElement* and can therefore be annotated with an arbitrary number of annotations. The top element *Corpus* contains a list of *Feature* objects, which represent names of annotations and annotation values represented by *FeatureValue*. A *Node* or *Edge* object contains a list of *Annotation* objects, having *Feature* and *FeatureValue* objects. This mechanism is analogous to the annotation mechanism of TigerXML. Since adding element typing has been a major point of interest in the development of <tiger2/>, we enhanced this mechanism by adding the attributes *Node.type* and *Edge.type*. To exemplify the power of the typing mechanism, we will consider the representation of dependency annotations, not originally supported in TigerXML, and the addition of a special class of nodes for compound items below. Figure 4 depicts a multilayer annotation for the syntactic fragment in (1):

(1)  *put up new wallpaper*

The different annotation layers are shown using visualisations from the ANNIS corpus search and visualisation system. The XML representation of this fragment is given in Figure 5 in the <tiger2/> format.

**Fig. 4** Graphic representation of a syntactic fragment annotated in multiple layers.

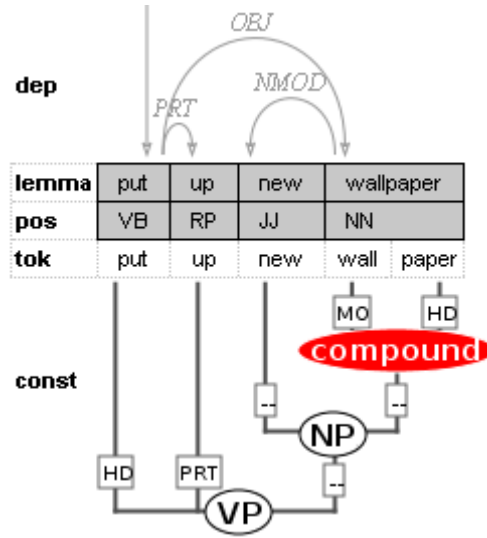

**Fig. 5** A syntactic fragment <s> showing (in bold) typed edges for dependencies, as well as typed terminals for compound stems and non-terminals for compound nodes.

```
<body>
<s xml:id="s1">
  <graph root="s1_ROOT">
    <terminals>
        <t              xml:id="s1_t1"            pos="VB"            lemma="put"
tiger2:corresp="tokens.xml#wordForm1">
          <edge tiger2:type="dep" tiger2:target="#s1_nt2" label="OBJ"/>
          <edge tiger2:type="dep" tiger2:target="#s1_t2" label="PRT"/>
        </t>
        <t              xml:id="s1_t2"                      pos="RP"            lemma="up"
tiger2:corresp="tokens.xml#wordForm2"/>
        <t              xml:id="s1_t3"              pos="JJ"            lemma="new"
tiger2:corresp="tokens.xml#wordForm3"/>
        <t              xml:id="s1_t4"              tiger2:type="stem"
tiger2:corresp="tokens.xml#wordForm4"/>
        <t              xml:id="s1_t5"              tiger2:type="stem"
tiger2:corresp="tokens.xml#wordForm5"/>
    </terminals>
    <nonterminals>
        <nt xml:id="s1_nt1" cat="VP">
          <edge tiger2:type="const" label="HD" tiger2:target="#s1_t1"/> <!-- put -->
          <edge tiger2:type="const" label="PRT" tiger2:target="#s1_t2"/> <!-- up -->
          <edge tiger2:type="const" label="DO" tiger2:target="#s1_nt2"/> <!-- NP -->
        </nt>
        <nt xml:id="s1_nt2" cat="NP">
          <edge tiger2:type="const" tiger2:target="#s1_t2"/> <!-- new -->
          <edge tiger2:type="const" tiger2:target="#s1_nt3"/><!--wallpaper-->
        </nt>
        <nt       xml:id="s1_nt2"        tiger2:type="compound"        pos="NN"
lemma="wallpaper">
```

```
            <edge tiger2:type="const" label="MO" tiger2:target="#s1_t4"/> <!-- wall- --
>
            <edge tiger2:type="const" label="HD" tiger2:target="#s1_t5"/> <!-- paper --
>
            <edge tiger2:type="dep" tiger2:target="#s1_t3" label="NMOD"/> <!- new
-->
```

The fragment is contained within a *Segment* element serialised by `<s>`.[2] The segment can contain one or multiple `<graph>`s, the latter possibility being used for instance in the case of alternative parses. In the `<terminals>` area, we see references to five terminals linked stand-off from an external file called tokens.xml. It is also possible to use inline terminals using the backwards compatible `@word` attribute from TigerXML. The non-terminals encompass both default syntactic constituents, which have been given no special type and are annotated with `@cat`, and a special compound node using the `@type` attribute.[3] The compound can have different annotations, such as `@pos` (part-of-speech) and `@lemma` annotations. Similarly, the compound stem terminals lack the lemma and pos annotations of other terminals in the current fragment.

We also see how *Edge* objects are typed as 'const' (for constituent trees) or 'dep' (for dependencies) much like *Node* objects can be typed as 'compound'. The values of the `@type` attribute are not specified in <tiger2/> and can be freely chosen for a specific corpus, though they must be declared in the `<annotation>` block in the `<head>` of the document as shown in Figure 6, and bound to an element domain (terminals `t`, non-terminals `nt`, or edge).

**Fig. 6** Example of an edge type definition for dependencies.

```
<head>
...
<annotation>
  <feature type="dep" domain="edge"/> <!-- declaration of edge type 'dep' -->
  <feature name="label" type="dep" domain="edge"> <!-- declaration of 'label'
annotation -->
    <value name="sbj">Subject</value> <!-- values for the 'label' annotation -->
  ...
  </feature>
…
```

Edges are defined using the `<feature>` element with the domain `edge`, analogous to the declaration of node features in TigerXML (the element `<edge>` is no longer used for the declaration). The edge type is specified as 'dep' for dependencies. It then becomes possible to define an annotation named 'label', which applies to edges of the type 'dep'. Some values for the annotation 'label' can be specified in the declaration of values, e.g 'sbj' for subject. This type of edges can then be used in the body of the document as in Figure 5 above.

---

[2] `<s>` stands for any syntactically annotated segment - this need not be a sentence, and can also be larger, as in a textual segment, or smaller as in a phrase.

[3] Reserved attributes like `@type` carry the namespace `tiger2` in order to allow a further user-defined attribute `@type`.

The definition of additional types of nodes works similarly, using the `@type` attribute. Figure 7 shows the declaration of the part-of-speech annotation for non-terminals (`domain="nt"`) of the type 'compound'. Note that in this manner compounds can have different allowed pos tags than other types of nodes, such as terminals, though features with open-ended attribute values can be defined as well, as in TigerXML.

**Fig. 7** Example of a node type definition, in this case for non-terminal compounds.

```
<feature type="compound" domain="nt"/>
<feature name="pos" type="compound" domain="nt"
dcr:datcat="http://www.isocat.org/datcat/
DC-396">
        <value name="JJ" dcr:datcat="http://www.isocat.org/datcat/DC-
1230">Adjective</value>
...
</feature>
```

Terminal nodes may also be used to represent 'empty' elements, such as unexpressed subjects in languages such as Italian or Spanish. For example, in the Italian sentence *Gliel'ho già dato* '(I) have already given it to him', the subject pronoun 'I' is not realized. At the same time, the clitic form *Gliel'*, shortened from *Glielo* contains two grammatical functions: indirect object 'to him' and direct object 'it'. This form is then written together with the auxiliary *ho* '(I) have', creating an orthographic unit. Figure 8 gives a possible representation of the dependencies in this sentence using <tiger2/>.

**Fig. 8** Example of an Italian sentence with an 'empty' pronoun and clitics.

```
<s xml:id="s1">
  <graph root="s1_ROOT">
    <terminals>
        <t xml:id="s1_t1" tiger2:corresp="tokens.xml#wordForm1"/> <!-- [pro] -->
        <t xml:id="s1_t2" tiger2:corresp="tokens.xml#wordForm2"/> <!-- Glie -->
        <t xml:id="s1_t3" tiger2:corresp="tokens.xml#wordForm3"/> <!-- l' -->
        <t xml:id="s1_t4" tiger2:corresp="tokens.xml#wordForm4"/> <!-- ho -->
        <t xml:id="s1_t5" tiger2:corresp="tokens.xml#wordForm5"/> <!-- già -->
        <t xml:id="s1_t6" tiger2:corresp="tokens.xml#wordForm6"> <!-- dato -->
         <edge tiger2:type="dep" tiger2:target="#s1_t1" label="SUBJ"/>
         <edge tiger2:type="dep" tiger2:target="#s1_t2" label="INDOBJ"/>
         <edge tiger2:type="dep" tiger2:target="#s1_t3" label="OBJ"/>
         <edge tiger2:type="dep" tiger2:target="#s1_t4" label="AUX"/>
         <edge tiger2:type="dep" tiger2:target="#s1_t5" label="ADV"/>
        </t>
    </terminals>
    <nonterminals>
        <nt xml:id="s1_nt1" tiger2:type="orthWord" orth="gliel'" lemma="glielo">
```

```
        …
        </nt>
        <nt xml:id="s1_nt2" tiger2:type="orthContraction" orth="gliel'ho">
        …
        </nt>
    </nonterminals>
        ...
```

The 'empty' subject pronoun [pro] is added as a subject for the sentence (in this analysis all arguments are governed by the participle *dato* 'given', not the auxiliary). The complex pronoun *gliel'* is split into two units for each grammatical function and then merged via a non-terminal of the type `orthWord`, which also gives the complete form of the word *glielo*. Finally, the further contraction with the auxiliary *ho* is represented by a non-terminal of the type `orthContraction`.

Which phenomena should be represented by what <tiger2/> objects and types (e.g. empty elements like traces, phrasal verbs like *put up*, subtokenisation phenomena) is a corpus design decision left up to individual data curators, though in either case, use of standardised semantics via ISOCat is recommended. For this purpose we provide the model attributes *Feature.dcrReference* and *FeatureValue.dcrReference* to carry a URI value pointing to an ISOCat entry. Their serialisation can also be seen in Figure 7, which specifies that the annotation named 'pos' refers to the ISOCat datapoint describing part-of-speech annotation, as well as a datapoint for a specific part-of-speech annotation for adjectives, both using the `@dcr:datcat` attribute.

A further possibility envisioned in <tiger2/> is the binding of other data resources, and especially of ISO standards, such as the Morpho-syntactic Annotation Framework MAF. Here we use the model attribute *Terminal.source*, which addresses a URI to another data source, for instance *wordForm* objects in the MAF meta-model which could then describe morphological phenomena such as compounding in an external file. In the XML representation, the terminal corresponding to *wallpaper* in Figure 5 could then look as follows using the `@corresp` attribute:

```
<t xml:id="s1_t4" corresp="MAF.xml#wordForm4"/> <!-- wallpaper -->
```

where the MAF document would also contain the relevant pos and lemma annotations, e.g. using MAF syntax:

```
<maf>
…
    <token xml:id="t4">wall</token>
    <token xml:id="t5">paper</token>
…
    <wordForm xml:id="wordForm4" lemma="wallpaper" tokens="t4 t5">
        <fs>
            <f name="pos"> <symbol value="NN"/> </f>
        </fs>
    </wordForm>
```

…
</maf>

For further examples using the typing mechanism, MAF integration and an exhaustive comparison of TigerXML and <tiger2/>, the interested reader is referred to the current documentation available on the <tiger2/> website: http://korpling.german.hu-berlin.de/tiger2/.

## 4.2 A Case in Point: TüBa-D/Z in <tiger2/>

As a first application of the concepts described above to a real use-case, we have converted the TüBa-D/Z Treebank (Telljohann et al. 2004, 2009) to <tiger2/>. The original treebank was coded in TigerXML and contained two types of annotated nodes: syntactic phrase categories and topological fields (used in German e.g. to distinguish pre- and post-verbal domains in main clauses and positions after the complementiser in subordinate clauses). Though these types of annotation are unrelated, the constraints of the original TigerXML format forced a nesting of syntactic categories within topological ones and vice versa, as illustrated in Figure 9 for the following sentence:

(2)    *"Es gab Teilnehmer, die Umsatzzahlen gelernt haben."*
       "There were participants who learned sales-numbers."

**Fig. 9** TigerXML tree mixing syntactic and topological nodes. There is no way to represent binding edges for the coreferent object and relative pronoun.

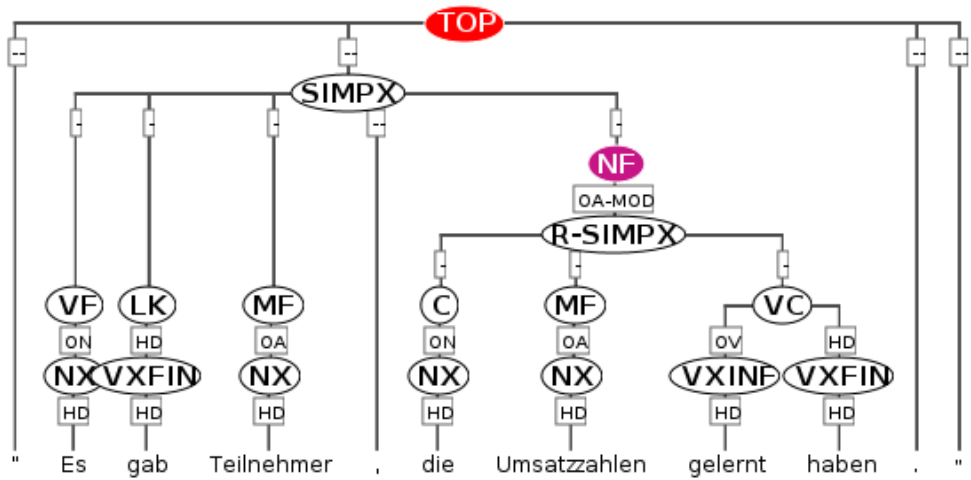

For example, the nominal phrase *Teilnehmer* (NX) is embedded within the middle field (MF), the domain after the main clause finite verb. It is difficult to search for the direct object of the sentence (the edge above NX with the label OA for 'object, accusative'), since corpus users cannot know if the object will be realised in the middle field MF or e.g. in the preverbal field (or Vorfeld, VF).

With the addition of coreference annotation to the treebank in version 5, the corpus now also contains multiple types of edges: syntactic dominance edges and various types of coreference edges, such as anaphoric binding between *Teilnehmer*

'Participants' and the relative pronoun *die* 'who'. These challenges make the corpus an ideal use-case for separating types of nodes and edges to make them explicitly queryable (see Krause et al. 2011 for more details). The result of the separation, which can easily be expressed in <tiger2/> is shown in Figure 10.

**Fig. 10** Two separate trees annotating the same tokens in one <tiger2/> document. Topological fields (below) do not interfere with constituents (above).

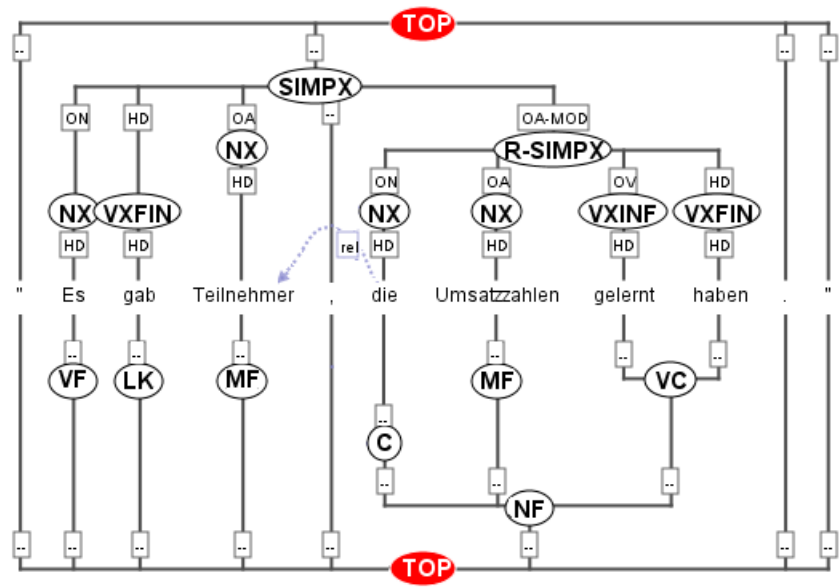

It is now easy to find the direct object of the sentence regardless of topological fields, and likewise to query topological fields independently of the syntactic phrases which encompass them. It is also possible to include the coreference edge missing in the TigerXML version. Finally, for backwards compatibility it is even possible to represent a third tree with all nodes from the original structure using a third, hybrid node type to retain the original tree representation in parallel to the other two trees, if one so wishes. Since the expressivity of the format is graph-like (more specifically, the API supports an unbounded set of directed acyclic graphs), any *n* trees can be combined in a <tiger2/> project.[4] An excerpt of the <tiger2/> XML code for the graph in Figure 10 is given in Figure 11.

---

[4] An anonymous reviewer has asked whether <tiger2/> data could also be represented in a tuple store (e.g. in RDF): this is certainly possible. It should be noted that not all conceivable graphs can be represented in the format. For example *n* to *m* edges connecting multiple nodes at a time are not supported.

**Fig. 11** XML representation of some nodes and edges from Figure 10.

```
<s xml:id="s1">
  <graph root="s1_ROOT">
    <terminals>
        ...
        <t xml:id="s1_t4" pos="NN" lemma="Teilnehmer"
tiger2:word="Teilnehmer"/>
        <t xml:id="s1_t5" pos="$," lemma="," tiger2:word=","/>
        <t xml:id="s1_t6" pos="PRELS" lemma="d" tiger2:word="die">
          <edge tiger2:type="coref" tiger2:target="#s1_t4" label="rel"/>
        </t>
        ...
    </terminals>
    <nonterminals>
        ...
        <nt xml:id="s1_nt_const3" cat="NX">
          <edge tiger2:type="const" label="HD" tiger2:target="#s1_t4"/> <!--
Teilnehmer -->
        </nt>
        <nt xml:id="s1_nt_const4" cat="NX">
          <edge tiger2:type="const" label="HD" tiger2:target="#s1_t6"/> <!-- die -->
        </nt>
        ...
        <nt xml:id="s1_nt_field3" tiger2:type="field" field="MF">
          <edge tiger2:type="field" tiger2:target="#s1_t4"/> <!-- Teilnehmer -->
        </nt>
        <nt xml:id="s1_nt_field4" tiger2:type="field" field="C">
          <edge tiger2:type="field" tiger2:target="#s1_t6"/> <!-- die -->
        </nt>
        ...
```

## 5 Conclusion

In the previous sections we have presented <tiger2/>, a new XML format and meta-model for syntactic annotation serialising the data-model defined by SynAF (ISO 24615:2010). By further developing a widely accepted existing XML format, our approach has been to change as little as possible and as much as necessary to remain as close as possible to the already disseminated best practices defined by TigerXML. <tiger2/> is capable of representing constituent and dependency structures, distinguishing an arbitrary number of types of edges and terminal or non-terminal nodes, which may carry different key-value annotations. This allows the format to express a variety of phenomena which have been partly discussed here, such as coreference, as well as many conceivable implementations not exemplified above, such as the use of empty or trace elements to represent transformational theories. Binding of external resources such as stand-off source tokens, morpho-syntactically annotated documents in MAF and external reference to the ISOCat Data Category Registry in the annotation declaration all allow further expressivity and interoperability for development with the SynAF data-model.

As a first major resource and test case in <tiger2/> we have briefly outlined the conversion of the multi-layer edition of the German TüBa-D/Z treebank described in Krause et al. (2011) to the new format. An XSD specification for <tiger2/> is already available for download from the <tiger2/> website (http://korpling.german.hu-berlin.de/tiger2/), along with examples representing various phenomena. Using a meta-model approach and the SaltNPepper converter framework (cf. Zipser & Romary 2010), we also offer Java-based converters for a variety of XML and non-XML formats to <tiger2/>. For the near future we are planning the public release of the Java API for <tiger2/> which will facilitate the conversion of further existing formats to <tiger2/> and allow programmatic manipulation of resources in the format.

**ISO Standards/Drafts:**